\documentclass{article}
\usepackage{ijcai24}

\usepackage{times}
\usepackage{soul}
\usepackage{url}
\usepackage[hidelinks]{hyperref}
\usepackage[utf8]{inputenc}
\usepackage[small]{caption}
\usepackage{graphicx}
\usepackage{amsmath}
\usepackage{amssymb}
\usepackage{amsthm}
\usepackage{booktabs}
\usepackage{algorithm}
\usepackage{algorithmic}
\usepackage[switch]{lineno}

\usepackage{arydshln}

\usepackage{multirow}
\usepackage[table,xcdraw]{xcolor}
\usepackage{bbding}
\usepackage{float}
\usepackage{verbatim}
\usepackage{color}

\title{Unsupervised Anomaly Detection via Masked Diffusion Posterior Sampling}

\author{
    Di Wu$^1$
\and
Shicai Fan$^{1,2}$
\and
Xue Zhou$^{1,2}$
\and
Li Yu$^{1,2}$
\and
Yuzhong Deng$^{1}$
\and
Jianxiao Zou$^{1,2}$
\and
Baihong Lin$^{2}$\footnote{B. Lin is the corresponding author. Source code will be available at \url{https://github.com/KevinBHLin/}.}
\affiliations
$^1$School of Automation Engineering, University of Electronic Science and Technology of China(UESTC)\\
$^2$Shenzhen Institute for Advanced Study, UESTC
\emails
{linbaihong111}@126.com
}

\begin{document}

\maketitle

\begin{abstract}
	Reconstruction-based methods have been commonly used for unsupervised anomaly detection, in which a normal image is reconstructed and compared with the given test image to detect and locate anomalies. Recently, diffusion models have shown promising applications for anomaly detection due to their powerful generative ability. However, these models lack strict mathematical support for normal image reconstruction and unexpectedly suffer from low reconstruction quality. To address these issues, this paper proposes a novel and highly-interpretable method named Masked Diffusion Posterior Sampling (MDPS). In MDPS, the problem of normal image reconstruction is mathematically modeled as multiple diffusion posterior sampling for normal images based on the devised masked noisy observation model and the diffusion-based normal image prior under Bayesian framework. Using a metric designed from pixel-level and perceptual-level perspectives, MDPS can effectively compute the difference map between each normal posterior sample and the given test image. Anomaly scores are obtained by averaging all difference maps for multiple posterior samples. Exhaustive experiments on MVTec and BTAD datasets demonstrate that MDPS can achieve state-of-the-art performance in normal image reconstruction quality as well as anomaly detection and localization.
\end{abstract}

\section{Introduction}

Anomaly detection (AD) is a fundamental computer vision task with widespread applications in medical diagnosis, intelligent manufacturing, autonomous driving and etc. However, due to the rarity and diversity of anomaly samples, recent studies mainly focus on unsupervised anomaly detection (UAD) \cite{diers2023survey}, 
in which the models only learn from normal samples but can detect anomaly data.

So far, there have existed various methods for UAD, among which reconstruction-based method is one of the earliest and most common neural network approaches \cite{ruff2021unifying}. Given a test image, reconstruction-based method tries to reconstruct the corresponding normal image and compute the difference between the test image and the reconstruction to detect and localize the anomalies. Obviously, how to reconstruct a normal image is a key issue for reconstruction-based methods. Early reconstruction models include Autoencoder (AE), Generative Adversarial Network (GAN) and etc. However, AE can easily suffer from ``identical shortcut'' \cite{you2022unified} and blurry reconstruction \cite{BAUR2021101952}, i.e., they reconstruct both normal and anomalous images and the results tend to be more blurry. Although GAN can alleviate the problems of AE, mode collapse and training instability make GAN challenging for UAD \cite{XIA2022497}.

Recently, Diffusion Models (DMs) have attracted most researchers' attention with their powerful image generation ability \cite{ho2020denoising}.
Compared with previous generative models, DMs can effectively record image priors and generate various realistic images after simple training based on Gaussian denoising. Thus, they show promising applications for UAD to alleviate the problems of inferior reconstruction quality or insufficient coverage of the normal image distribution.
However, two key problems arise when introducing DMs for UAD: 
First, although different DM-based methods are proposed for UAD, they are lack of strict mathematical theories or interpretability to ensure that the anomaly region of a test image can be reconstructed as the normal one. 
Second, most DMs unexpectedly suffer from low reconstruction quality, especially for the normal region of a test image, since Gaussian noise in DMs will destroy the original normal texture. This problem can easily result in misjudgment of anomalous pixels in the normal region. Thus, it requires further studies for DMs to maintain the texture in the normal region of a test image after reconstruction.

In this paper, we propose a novel Masked Diffusion Posterior Sampling (MDPS) method for UAD under Bayesian framework, which has high interpretability supported by relatively-strict mathematics. 
In our method, we firstly propose a Masked Noisy Observation Model which regards a test image as a masked noisy observation of a normal image to protect the normal region of a test image and enhance reconstruction quality.
Then, we take a Denoising Diffusion Implicit Model (DDIM) trained by normal samples as image prior, and model the problem of normal image reconstruction as multiple diffusion posterior samplings for normal images based on the devised observation model and the DDIM-based prior.
Third, using a metric designed from pixel-level and perceptual-level perspectives, we compute the difference maps between multiple normal posterior samples and the test image respectively, and average all difference maps to accurately obtain anomaly scores. 
Exhaustive experiments on MVTec and BTAD datasets prove that the proposed MDPS achieves excellent reconstruction quality and high accuracy of anomaly detection and localization compared with other state-of-the-art reconstruction-based methods.

\section{Related Work}
Recent reconstruction-based UAD methods can be roughly divided into three categories, including AE-based methods, GAN-based methods, and DM-based methods.

\paragraph{AE-based methods.} 
Early reconstruction-based UAD methods commonly adopt AE due to their simple architectures and easy-to-implement training processes. Unfortunately, AEs can easily suffer from the problems of ``identical shortcut''  and blurry reconstruction. For the first problem, since Vision Transformer (ViT) can prevent ``identical shortcut'' \cite{you2022unified}, recent studies try to design ViT-based AEs, e.g.,  \cite{mishra2021vt}, \cite{you2022adtr}, \cite{you2022unified} and etc. For the second problem, various methods have been proposed: \cite{bergmann2018improving} designs a loss function of structure similarity to improve reconstruction quality; \cite{liu2020towards} and \cite{dehaene2020iterative} introduce Variational AE to achieve better reconstruction results; \cite{zavrtanik2021draem} and \cite{SunWang-376} propose a self-supervised model to reduce the influence of blurry reconstruction. 
Nevertheless, the above schemes cannot totally avoid blurry reconstruction, thus, leading to performance bottleneck for UAD.

\paragraph{GAN-based methods.} 
To overcome the drawbacks of AEs, \cite{schlegl2017unsupervised} firstly proposed a GAN-based method named AnoGAN. Since then, various variants are proposed, e.g., \cite{akccay2019skip}, \cite{schlegl2019f} and etc. Although GANs can empirically generate high definition results, two problems make GANs challenging for UAD \cite{XIA2022497}: 
First, GAN training is highly unstable to converge, sometimes bringing in meaningless reconstruction. Since the training process of AE is more stable, AE and GAN are usually combined to alleviate training instability, e.g., \cite{tang2020anomaly}, \cite{CONTRERASCRUZ2023108470}. 
Second, due to the vanish of discriminator gradient, it is hard to ensure sufficient coverage of the normal distribution during GAN training process. This phenomenon is called mode collapse, which can easily lead to reconstruction results with only a few modes. To alleviate this problem, self-supervised models are proposed based on simulated anomaly data to guide GAN training, e.g., \cite{song2021anoseg}, \cite{LIU2023120284} and etc.

\paragraph{DM-based methods.}
Recently, DMs have been popular for academic research due to their potentially-powerful generative ability. Typical DMs include DDPM \cite{ho2020denoising}, DDIM \cite{song2021denoising}, SDE \cite{song2021scorebased}, etc. Compared with AEs and GANs, DMs have desirable properties, such as distribution coverage, a stationary training objective and easy scalability \cite{dhariwal2021diffusion}. To generate desirable images based on DMs for certain requirements, conditional DMs are studied, e.g., \cite{dhariwal2021diffusion} proposes a classifier-guided DM to generate images given a certain class label; \cite{chung2022diffusion} extends diffusion solvers to efficiently handle general noisy inverse problems via approximation of posterior sampling. The above works show promising applications for UAD to reconstruct high definition normal images.

So far, there have been a few DM-based AD methods. \cite{wolleb2022diffusion} and \cite{Pinaya2022} firstly introduce DMs for weakly supervised AD and UAD respectively in medical diagnosis. \cite{wyatt2022anoddpm} proposes a DM-based UAD method named AnoDDPM, which uses simplex noise to improve normal image reconstruction quality for brain MRI. \cite{gonzalez2023sano} uses score-based models to produce a gradient map highlighting anomaly areas. \cite{bercea2023mask} proposes a DM-based model named AutoDDPM which consists of three stages, i.e, mask, stitch, and re-sampling. \cite{lu2023removing} models the problem of normal image reconstruction as a DM-based denoising process. The above methods have enhanced the accuracy of DM-based AD from different aspects. However, these DM-based methods are lack of strictly-mathematical support, and unexpectedly suffer from low normal image reconstruction quality, especially for the normal region of a test image.

\section{Background}
In this section, we briefly review SDE-based diffusion models \cite{song2021scorebased} and DDIM \cite{song2021denoising}.
\paragraph{SDE-based diffusion models.}
For a diffusion process $\{\boldsymbol x(t)\}^T_{t=0}$ indexed by a continues time variable $t\in[0,T]$, let $\boldsymbol x(0)\sim p_0(x)$ and $\boldsymbol x(T)\sim p_T(\boldsymbol x)$, in which $p_0(\boldsymbol x)$ and $p_T(\boldsymbol x)$ denotes the data distribution of interest and a known spherical Gaussian distribution respectively. Then, the forward noising process $\boldsymbol x(0)\rightarrow \boldsymbol x(T)$ can be modeled as the following It\^o stochastic differential equation (SDE):
\begin{equation}
	\label{eq:forwardSDE}
	d\boldsymbol x = \boldsymbol f(\boldsymbol x,t)dt + g(t)d\boldsymbol \omega 
\end{equation}
where $\boldsymbol f(\cdot ,t)$ and $g(\cdot)$ are the drift and diffusion coefficient of $\boldsymbol x(t)$ respectively, $\boldsymbol \omega$ is a standard Wiener process. The corresponding reverse process $\boldsymbol x(T)\rightarrow \boldsymbol x(0)$ is given by 
\begin{equation}
	\label{eq:reverseSDE}
	d\boldsymbol x = [\boldsymbol f(\boldsymbol x,t)-g(t)^2 \nabla_{\boldsymbol x} \log p_t(\boldsymbol x)]dt + \boldsymbol g(t)d\hat{\boldsymbol \omega} ,
\end{equation}
where $\hat{\boldsymbol \omega}$ is a standard Wiener process running backwards, and $dt$ is an infinitesimal negative time step. $\nabla_{\boldsymbol x}\log p_t(\boldsymbol x)$ is the score function of each marginal distribution which can be approximately estimated by training a time-dependent score-based model $s_\theta(\boldsymbol x,t)$, i.e., $\boldsymbol s_\theta(\boldsymbol x,t)\simeq\nabla_{\boldsymbol x}\log p_t(\boldsymbol x)$.

\begin{figure*}[tp]
	\centering
	\setlength{\abovecaptionskip}{5pt}
	\includegraphics[width=1\linewidth]{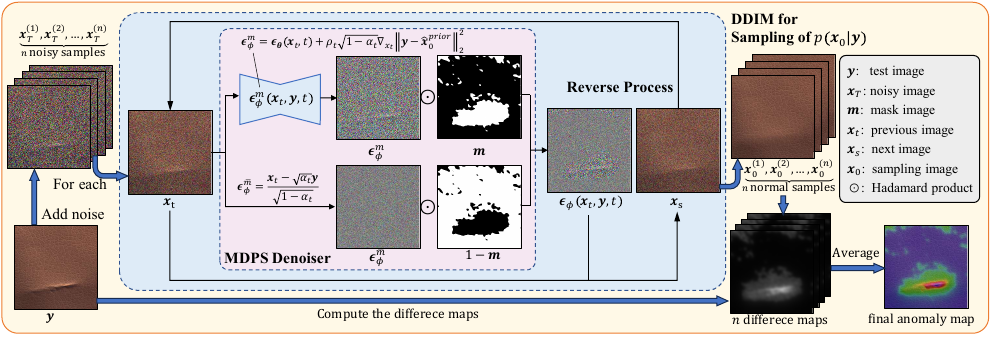}
	\caption{ 
		Overview of MDPS. The MDPS denoiser shown in the pink box is designed partially based on the denoiser $\boldsymbol{\epsilon}(\boldsymbol{x}_t,t)$ of DDIM for sampling of $p(x_0)$. Based on MDPS, we can obtain $n$ normal posterior samples from $n$ noisy versions of the test image $\boldsymbol y$ respectively. Then, the final anomaly map is obtained by averaging $n$ difference maps computed from $n$ normal posterior samples and the test image $\boldsymbol y$. 
	}
	\label{fig:method}
\end{figure*}

\paragraph{DDIMs.}
DDIMs can be regarded as an acceleration version of DDPMs \cite{ho2020denoising} with the same training procedure.
However, different from DDPMs, the forward noising process of DDIM is defined as a non-Markov process:
\begin{equation}
	\label{eq:forwardDDIM}
	\boldsymbol x_t = \sqrt{{\alpha}_t}\boldsymbol x_0+\sqrt{1-{\alpha}_t}\boldsymbol \epsilon, \boldsymbol \epsilon \sim \mathcal{N}(0, \mathbf{I})
\end{equation}
where $\{\alpha_t\in (0,1]| t \in [0,T]\}$ is a decreasing sequence decided by a predetermined schedule, and $\boldsymbol \epsilon$ is white noise drawn from a standard normal distribution.
The accelerated reverse process of DDIM is described as:
\begin{equation}
	\begin{split}
		\boldsymbol x_s=&\sqrt\frac{{\alpha}_s}{{\alpha}_t}(\boldsymbol x_t-\sqrt{1-{\alpha}_t} \boldsymbol \epsilon_{\theta}(\boldsymbol x_t,t)) \\&+\sqrt{1- {\alpha}_s - \sigma_t^2}\boldsymbol \epsilon_{\theta}(x_t,t) + \sigma_t \boldsymbol \epsilon_t,~~\boldsymbol \epsilon_t\sim \mathcal{N}(0, \mathbf{I})
	\end{split}
	\label{eq:reverseDDIM}
\end{equation}
where $\sigma_t=\sqrt{(1-{\alpha}_s)/(1-{\alpha}_t)} \sqrt{1-{\alpha}_t/{\alpha}_s}$, $s<t$, and $\boldsymbol \epsilon_{\theta}(x_t,t)$ is a U-net denoiser which estimates the white noise $\boldsymbol \epsilon$ from $\boldsymbol x_t$, and can be trained by minimizing the following objective:
\begin{equation}
	L(\theta) := \mathbb{E}_{t,\boldsymbol x_0,\boldsymbol \epsilon} [
	\Vert \boldsymbol \epsilon - \boldsymbol \epsilon_{\theta}(\sqrt{{\alpha}_t}\boldsymbol x_0+\sqrt{1-{\alpha}_t}\boldsymbol \epsilon, t)\Vert^2]
	\label{eq:DDIM_loss}
\end{equation}
\cite{song2021scorebased} has pointed out that DDIM is a discrete form of SDE when $\boldsymbol f(x,t) = -\boldsymbol x_t\beta(t)/2$, $g(t)=\sqrt{\beta(t)}$, and $\alpha_t=\prod_t{(1-\beta(t))}$ in Eqn.(\ref{eq:forwardSDE}). Thus, 
\begin{equation}
	\begin{split}
		\boldsymbol \epsilon_{\theta}(x_t,t)&= -\sqrt{1-\alpha_t}\boldsymbol s_\theta(x,t) \\
		&\simeq -\sqrt{1-\alpha_t}\nabla_{\boldsymbol x_t}\log p_t(\boldsymbol x_t)
	\end{split}
	\label{eq:DDIMandSDE}
\end{equation}

\section{Masked Diffusion Posterior Sampling}
In this section, we firstly establish a masked noisy observation model for normal image reconstruction. Then, based on the observation model, we propose a posterior sampling method for normal images using DDIMs. Thirdly, we design image-wise and pixel-wise anomaly scores and propose a mask generation scheme based on the designed anomaly scoring for the proposed observation model to enhance the reconstruction quality of posterior samples.

\subsection{Masked Noisy Observation Model}
Let $\boldsymbol{y}$ and $\boldsymbol{x}_0$ denote an anomaly image and the corresponding normal image respectively. Let $\boldsymbol m$ denote a mask image with the same size of $\boldsymbol y$ in which pixel values are set to 0 in the normal regions of $\boldsymbol{y}$ and set to 1 otherwise. Then, we establish a masked noisy observation model to describe the relationship between $\boldsymbol{y}$ and $\boldsymbol{x}_0$ as the following:
\begin{equation}\label{eq:observation}
	\boldsymbol{y}  =(1- \boldsymbol{m}) \odot \boldsymbol{x}_0 + \boldsymbol{m} \odot (\boldsymbol{x}_0 + \boldsymbol{n})
\end{equation}
where $\odot$ denotes Hadamard product, $\boldsymbol{n}$ denotes zero-mean isotropic Gaussian noise and $\boldsymbol{n} \sim \mathcal{N}(0, \sigma ^2 \boldsymbol{I})$. Obviously, the above modeling assumes that $\boldsymbol{y}$ is a noisy observation of $\boldsymbol{x}_0$ in the anomaly region whereas $\boldsymbol{y}$ shares the same pixel values as that of $\boldsymbol{x}_0$ in the normal region, i.e.,
\begin{subequations}
	\begin{align}
		(1- \boldsymbol{m}) \odot \boldsymbol{y} &= (1- \bold m) \odot \boldsymbol{x}_0 \label{eq:normal} \\
		\boldsymbol m \odot \boldsymbol{y} &=\boldsymbol m \odot (\boldsymbol{x}_0 + \boldsymbol{n}). \label{eq:abnormal}
	\end{align}
\end{subequations}
As $\boldsymbol{y}$ would contain unknown types of anomaly, the assumption of Gaussian noise for anomaly is reasonable. Furthermore, the mask $1-\boldsymbol{m}$ can guarantee the reconstruction quality of normal image $\boldsymbol{x}_0$, since it maintains pixel values in the normal region of $\boldsymbol{y}$ for reconstruction of $\boldsymbol{x}_0$. Then, based on Eqn.(\ref{eq:observation}), we can obtain the distribution of $\boldsymbol{y}$ given $\boldsymbol{x}_0$ for the anomaly region indicated by $\boldsymbol{m}$ as the following:
\begin{equation}
	p\left(\boldsymbol{y}|\boldsymbol{x}_0\right)\sim \mathcal{N}(\boldsymbol{x}_0, \sigma^2\mathbf{I})
	\label{eq:observationdis}
\end{equation}

\subsection{Posterior Sampling for Normal Images}

To obtain multiple normal samples of $\boldsymbol{x}_0$  given $\boldsymbol{y}$, we propose to conduct posterior sampling for $p(\boldsymbol{x}_0|\boldsymbol{y})$. According to Bayes' Theorem, $p(\boldsymbol{x}_0|\boldsymbol{y})$ is related to the normal image prior $p(\boldsymbol{x}_0)$ and the observation distribution $p(\boldsymbol{y}|\boldsymbol{x}_0)$ given in Eqn.(\ref{eq:observationdis}). In our model, we introduce a DDIM trained by normal samples to model the normal image prior $p(\boldsymbol{x}_0)$, since this model has been widely adopted for its powerful image generation ability in recent years. Then, following \cite{chung2022diffusion}, we regard the posterior sampling process as a diffusion process based on the introduced DDIM for $p(\boldsymbol{x}_0)$, and obtain multiple normal samples drawn from $p(\boldsymbol{x}_0|\boldsymbol{y})$. The details are shown as the following. 

Specifically, let $\boldsymbol\epsilon_{\theta}(\boldsymbol{x}_t,t)$ and  $\boldsymbol\epsilon_{\phi}(\boldsymbol{x}_t,\boldsymbol{y},t)$ denote the denoisers of two DDIMs for sampling of $p(\boldsymbol{x}_0)$ and $p(\boldsymbol{x}_0|\boldsymbol{y})$ respectively. Since $\boldsymbol\epsilon_{\theta}(\boldsymbol{x}_t,t)$ is known and trained by normal images, we discuss the design of $\boldsymbol\epsilon_{\phi}(\boldsymbol{x}_t,\boldsymbol{y},t)$. 
According to Eqn.(\ref{eq:forwardDDIM}) and Eqn.(\ref{eq:normal}), we can accurately estimate $\boldsymbol\epsilon$ as: 
\begin{equation}\label{eq:normalPosteriorDenoiser}
	\boldsymbol\epsilon_{\phi}^{\boldsymbol{\overline{m}}}(\boldsymbol{x}_t,\boldsymbol{y},t)=\frac{\boldsymbol{x}_t - \sqrt{{\alpha}_t}\boldsymbol{x}_0}{\sqrt{{1-\alpha}_t}}=\frac{\boldsymbol{x}_t - \sqrt{{\alpha}_t}\boldsymbol{y}}{\sqrt{{1-\alpha}_t}}
\end{equation}
where $\boldsymbol\epsilon_{\phi}^{\boldsymbol{\overline{m}}}(\boldsymbol{x}_t,\boldsymbol{y},t)$ denotes the values of $\boldsymbol\epsilon_{\phi}(\boldsymbol{x}_t,\boldsymbol{y},t)$ in the normal region indicated by $1-\boldsymbol{m}$.

However, for the anomaly region indicated by $\boldsymbol{m}$, according to Bayes' Theorem, Eqn.(\ref{eq:DDIMandSDE}) and Eqn.(\ref{eq:abnormal}),
\begin{equation}\label{eq:anomalyPosteriorDenoiser}
	\begin{split}
		\boldsymbol\epsilon_{\phi}^{\boldsymbol{m}}(&\boldsymbol{x}_t,\boldsymbol{y},t)\simeq-\sqrt{1-\alpha_t}\nabla_{\boldsymbol{x}_t}\log p_t(\boldsymbol{x}_t|\boldsymbol{y}) \\
		&= -\sqrt{1-\alpha_t}(\nabla_{\boldsymbol{x}_t}\log {p(\boldsymbol{x}_t)}+\nabla_{\boldsymbol{x}_t}\log {p(\boldsymbol{y}|\boldsymbol{x}_t)})\\
		&\simeq\boldsymbol\epsilon_{\theta}(\boldsymbol{x}_t,t) - \sqrt{1-\alpha_t}\nabla_{\boldsymbol{x}_t}\log {p(\boldsymbol{y}|\boldsymbol{x}_t)}
	\end{split}
\end{equation}
where $\boldsymbol\epsilon_{\phi}^{\boldsymbol{m}}(\boldsymbol{x}_t,\boldsymbol{y},t)$ denotes the result of $\boldsymbol\epsilon_{\phi}(\boldsymbol{x}_t,\boldsymbol{y},t)$ for the anomaly region indicated by $\boldsymbol{m}$. Obviously, $\boldsymbol\epsilon_{\phi}^{\boldsymbol{m}}(\boldsymbol{x}_t,\boldsymbol{y},t)$ can be replaced by the above equation base on $\boldsymbol\epsilon_{\theta}(\boldsymbol{x}_t,t)$ and $\nabla_{\boldsymbol{x}_t}\log {p(\boldsymbol{y}|\boldsymbol{x}_t)}$. Since $\boldsymbol\epsilon_{\theta}(\boldsymbol{x}_t,t)$ is a known denoiser trained by normal samples, we focus on the modeling of $p(\boldsymbol{y}|\boldsymbol{x}_t)$.

As has been proven in \cite{chung2022diffusion}, 
\begin{equation}
	p(\boldsymbol{y}|\boldsymbol{x}_t) 
	= \int p(\boldsymbol{y}|\boldsymbol{x}_0)p(\boldsymbol{x}_0|\boldsymbol{x}_t) d\boldsymbol{x}_0 
	\label{eq:int_com}
\end{equation}
where $p(\boldsymbol{y}|\boldsymbol{x}_0)$ has been given in Eqn.(\ref{eq:observationdis}), and $p(\boldsymbol{x}_0|\boldsymbol{x}_t)$ is determined by the reverse process of DDIMs for sampling of $p(\boldsymbol{x}_0)$. Based on Eqn.(\ref{eq:forwardDDIM}) and Eqn.(\ref{eq:reverseDDIM}), 
\begin{equation}
	\hat{\boldsymbol{x}}_0^{\textit{prior}} = \frac{1}{\sqrt{{\alpha}_t}}(\boldsymbol{x}_t-\sqrt{1-{\alpha}_t} \boldsymbol{\epsilon}_{\theta}(\boldsymbol{x}_t,t)).
	\label{eq:x0}
\end{equation}
Apparently, in the reverse process of DDIMs, $\boldsymbol\epsilon$ of Eqn.(\ref{eq:forwardDDIM}) turns to be an unknown fixed constant, and can be estimated by $\boldsymbol\epsilon_{\theta}(\boldsymbol{x}_t,t)$. The estimation errors between $\boldsymbol\epsilon$ and $\boldsymbol\epsilon_{\theta}(\boldsymbol{x}_t,t)$ are reduced by minimizing the objective function of Eqn.(\ref{eq:DDIM_loss}) during the training process of DDIMs for sampling of $p(\boldsymbol{x}_0)$. Thus, we can assume that $\boldsymbol\epsilon\sim \mathcal{N}(\boldsymbol\epsilon_{\theta}(\boldsymbol{x}_t,t), \lambda^{-1}_t\mathbf{I})$ in which $\lambda_t$ indicates the estimation precision, and obtain
\begin{equation}
	p(\boldsymbol{x}_0|\boldsymbol{x}_t)\sim \mathcal{N}(\hat{\boldsymbol{x}}_0^{\textit{prior}}, {\frac{1-{\alpha}_t}{{\lambda_t\alpha}_t}}\mathbf{I})
	\label{eq:x0_dis}
\end{equation}
Then, for the anomaly region indicated by $\boldsymbol{m}$, $p(\boldsymbol{y}|\boldsymbol{x}_t)$  can be calculated based on Eqn.(\ref{eq:observationdis}), Eqn.(\ref{eq:int_com}) and Eqn.(\ref{eq:x0_dis}) as:
\begin{equation}
	p(\boldsymbol{y}|\boldsymbol{x}_t)\sim \mathcal{N}\left(\hat{\boldsymbol{x}}_0^{\textit{prior}}, \rho_t^{-1}\mathbf{I}\right),~\rho_t^{-1}=\frac{1-\alpha_t}{\lambda_t\alpha_t}+\sigma^2
	\label{eq:y_xt_dis}
\end{equation}
For simplicity, we let $\rho_t=\rho$, $\rho$ is an adjustable guidance scale. $\boldsymbol\epsilon_{\phi}^{\boldsymbol{m}}(\boldsymbol{x}_t,\boldsymbol{y},t)$ can be obtained based on Eqn.(\ref{eq:y_xt_dis}) as:
\begin{equation}\label{eq:anomalyPosteriorDenoiser2}
	\boldsymbol\epsilon_{\phi}^{\boldsymbol{m}}(\boldsymbol{x}_t,\boldsymbol{y},t)=\boldsymbol\epsilon_{\theta}(\boldsymbol{x}_t,t) + \rho \sqrt{1-{\alpha}_t} \nabla_{\boldsymbol x_t} \Vert \boldsymbol y- \hat{\boldsymbol{x}}_0^{\textit{prior}}\Vert_2^2
\end{equation}

Combining Eqn.(\ref{eq:anomalyPosteriorDenoiser2}) and Eqn.(\ref{eq:normalPosteriorDenoiser}), we can obtain the denoiser $\boldsymbol\epsilon_{\phi}(\boldsymbol{x}_t,\boldsymbol{y},t)$ of DDIMs for sampling of $p(\boldsymbol{x}_0|\boldsymbol{y})$ as:
\begin{equation}\label{eq:posteriorDenoiser}
	\begin{split}
		\boldsymbol\epsilon_{\phi}&(\boldsymbol{x}_t,\boldsymbol{y},t)= (1-\boldsymbol{m})\odot\frac{\boldsymbol{x}_t - \sqrt{{\alpha}_t}\boldsymbol{y}}{\sqrt{{1-\alpha}_t}} \\
		&+ \boldsymbol{m}\odot(\boldsymbol\epsilon_{\theta}(\boldsymbol{x}_t,t) + \rho \sqrt{1-{\alpha}_t} \nabla_{\boldsymbol x_t} \Vert \boldsymbol y- \hat{\boldsymbol{x}}_0^{\textit{prior}} \Vert_2^2)
	\end{split}
\end{equation}

\begin{algorithm}[t]
	\caption{Masked Diffusion Posterior Sampling}
	\textbf{Input}: test image $\boldsymbol y$, DDIM denoiser $\boldsymbol \epsilon_{\theta}(\boldsymbol x_t,t)$, mask image $\boldsymbol m$, guidance scale $\rho$, noise level $T$,  sampling times $N$\\
	\textbf{Output}: normal image $\boldsymbol x_0$
	\begin{algorithmic}[1] 
		\STATE $\boldsymbol \epsilon \sim \mathcal{N}(0, \mathbf{I})$.
		\STATE $\boldsymbol x_T = \sqrt{{\alpha}_T}\boldsymbol y + \sqrt{1-{\alpha}_T}\boldsymbol\epsilon$.
		\FOR{ \textbf{all} $n$ from $N$ to 1 }
		\STATE $t=\frac{Tn}{N}$, $s=\frac{T(n-1)}{N}$
		\STATE $\hat{\boldsymbol{x}}_0^{\textit{prior}} = \frac{1}{\sqrt{{\alpha}_t}}(\boldsymbol x_t-\sqrt{1-{\alpha}_t} \boldsymbol\epsilon_{\theta}(\boldsymbol x_t,t))$
		\STATE $\boldsymbol\epsilon_{\phi}^{\boldsymbol{m}} = \boldsymbol\epsilon_{\theta}(\boldsymbol{x}_t,t) + \rho \sqrt{1-{\alpha}_t} \nabla_{\boldsymbol x_t} \Vert \boldsymbol y- \hat{\boldsymbol{x}}_0^{\textit{prior}} \Vert_2^2$
		\STATE $\boldsymbol\epsilon_{\phi}^{\boldsymbol{\overline{m}}} = \frac{\boldsymbol{x}_t - \sqrt{{\alpha}_t}\boldsymbol{y}}{\sqrt{{1-\alpha}_t}}$
		\STATE $\boldsymbol\epsilon_{\phi}= (1-\boldsymbol{m})\odot\epsilon_{\phi}^{\boldsymbol{\overline{m}}}
		+ \boldsymbol{m}\odot\epsilon_{\phi}^{\boldsymbol{m}}$
		\STATE $\hat{\boldsymbol x}_0 = \frac{1}{\sqrt{{\alpha}_t}}(\boldsymbol x_t-\sqrt{1-{\alpha}_t} \boldsymbol \epsilon_{\phi}) $
		\STATE $\boldsymbol\epsilon_t \sim \mathcal{N}(0, \mathbf{I})$
		\STATE $\boldsymbol x_s=\sqrt{{\alpha}_s}\hat{\boldsymbol x}_0 + 
		\sqrt{1- {\alpha}_s - \sigma_t^2}\boldsymbol \epsilon_{\phi} +
		\sigma_t \boldsymbol \epsilon_t$
		\ENDFOR
		\STATE \textbf{return} $\boldsymbol x_0$
	\end{algorithmic}
	\label{alg:MDPS}
\end{algorithm}

\begin{table*}[t]
	\centering
	\setlength{\abovecaptionskip}{0cm}
	\renewcommand\arraystretch{1.1}
	\tabcolsep=0.04cm
	\newcommand{\tabincell}[2]{\begin{tabular}{@{}#1@{}}#2\end{tabular}}
	\caption{Anomaly detection and localization performance on MVTec. (Image-AUROC \%, Pixel-AUROC \%)}
	\begin{tabular}{cccccccccccc}
		\hline
		\multirow{2}{*}{Method} & \multicolumn{2}{c}{AE-based Methods}    &  & \multicolumn{2}{c}{GAN-based Methods}   &  & \multicolumn{5}{c}{DM-based Methods}                                 \\ \cline{2-3} \cline{5-6} \cline{8-12} 
		& DRÆM        & UniAD       &  & AnoGAN      & AnoSeg      &  & AnoDDPM     & AutoDDPM    & RAN         & \tabincell{c}{\textbf{MDPS}\\($N_s=1$)}        & \tabincell{c}{\textbf{MDPS}\\($N_s=16$)}   \\ \hline
		Carpet                  & (97.0, 95.5) & (99.8, 98.5) &  & (33.7, 54.0) & (96.0, \textbf{99.0}) &  & (54.6, 63.8) & (85.9, 86.0) & (\textbf{99.9}, 98.9) & (98.7, 93.4) & (99.6, 94.4) \\
		Grid                    & (99.9, \textbf{99.7}) & (98.2, 96.5) &  & (87.1, 58.0) & (99.0, 99.0) &  & (96.0, 82.3) & (\textbf{100}, 97.5)  & (99.7, 99.1) & (\textbf{100}, 99.4)  & (\textbf{100}, 99.4)  \\
		Leather                 & (\textbf{100}, 98.6)  & (\textbf{100}, 98.8)  &  & (45.1, 64.0) & (99.0, 98.0) &  & (96.9, 85.6) & (90.0, 91.5) & (\textbf{100}, \textbf{99.5})  & (\textbf{100}, 99.4)  & (\textbf{100}, 99.5)  \\
		Tile                    & (99.6, \textbf{99.2}) & (99.3, 91.8) &  & (40.1, 50.0) & (98.0, 98.0) &  & (96.5, 76.6) & (86.8, 74.0) & (98.0, 92.1) & (99.8, 95.4) & (\textbf{100}, 96.4)  \\
		Wood                    & (\textbf{99.1}, 96.4) & (98.6, 93.2) &  & (56.7, 62.0) & (99.0, \textbf{98.0}) &  & (87.2, 68.5) & (98.5, 80.5) & (98.1, 94.5) & (99.0, 94.1) & (\textbf{99.1}, 95.7) \\
		Bottle                  & (99.2, \textbf{99.1}) & (99.7, 98.1) &  & (80.0, 86.0) & (98.0, 99.0) &  & (87.8, 68.9) & (99.0, 97.4) & (99.3, 97.7) & (\textbf{100}, 98.7)  & (\textbf{100}, 98.6)  \\
		Cable                   & (91.8, 94.7) & (95.2, 97.3) &  & (47.7, 78.0) & (98.0, \textbf{99.0}) &  & (71.6, 64.1) & (84.7, 88.3) & (91.2, 95.6) & (97.1, 95.9) & (\textbf{98.3}, 95.8) \\
		Capsule                 & (\textbf{98.5}, 94.3) & (86.9, \textbf{98.5}) &  & (44.2, 84.0) & (84.0, 90.0) &  & (60.1, 72.5) & (61.8, 92.8) & (84.1, 97.5) & (91.7, 93.2) & (91.4, 93.6) \\
		Hazelnut                & (\textbf{100}, \textbf{99.7})  & (99.8, 98.1) &  & (25.9, 87.0) & (98.0, 99.0) &  & (69.5, 75.5) & (96.5, 92.9) & (97.9, 97.3) & (99.6, 98.6) & (99.8, 98.6) \\
		Metalnut                & (98.7, \textbf{99.5}) & (99.2, 94.8) &  & (28.4, 76.0) & (95.0, 99.0) &  & (55.5, 76.0) & (93.5, 94.8) & (99.2, 96.8) & (\textbf{100}, 97.3)  & (99.9, 97.7) \\
		Pill                    & (\textbf{98.9}, 97.6) & (93.7, 95.0) &  & (71.1, 87.0) & (87.0, 94.0) &  & (75.7, 73.6) & (59.3, 92.2) & (64.7, 92.5) & (96.2, 99.0) & (96.8, \textbf{99.2}) \\
		Screw                   & (93.9, 97.6) & (87.5, 98.3) &  & (10.0, 80.0) & (\textbf{97.0}, 91.0) &  & (64.7, 79.1) & (77.5, 93.0) & (89.9, \textbf{99.0}) & (93.1, 98.7) & (96.7, 98.9) \\
		Toothbrush              & (\textbf{100}, 98.1)  & (94.2, 98.4) &  & (43.9, 90.0) & (99.0, 96.0) &  & (57.2, 87.8) & (92.5, 96.9) & (96.9, 98.6) & (\textbf{100}, 98.7)  & (\textbf{100}, \textbf{98.8})  \\
		Transistor              & (93.1, 90.9) & (99.8, \textbf{97.9}) &  & (69.2, 80.0) & (96.0, 96.0) &  & (70.8, 63.0) & (80.1, 77.8) & (92.3, 93.1) & (99.9, 94.1) & (\textbf{100}, 94.7)  \\
		Zipper                  & (100, \textbf{98.8})  & (95.8, 96.8) &  & (71.5, 78.0) & (99.0, 98.0) &  & (92.0, 66.5) & (96.3, 89.0) & (85.5, 97.6) & (\textbf{100}, 98.5)  & (\textbf{100}, 98.5)  \\ \hline
		Average                     & (98.0, \textbf{97.3}) & (96.5, 96.8) &  & (50.3, 74.3) & (96.1, 96.9) &  & (75.7, 73.6) & (86.8, 89.6) & (93.1, 96.7) & (98.4, 97.0) & (\textbf{98.8}, \textbf{97.3}) \\ \hline
	\end{tabular}
	\label{MVTec_table}
	\vspace{-0.2cm}
\end{table*}

\subsection{Anomaly Scoring and Mask Generation}

To detect and locate anomaly, we have to predict image-wise and pixel-wise anomaly scores respectively. For prediction of pixel-wise anomaly score, we have to calculate the difference between the test image and the corresponding normal reconstruction result. Traditionally, pixel-level metrics such as $\mathcal{L}_1$-norm, and MSE are adopted to measure the differences of two images. Recently, \cite{zhang2018unreasonable} proposed a perceptual-level metric named LPIPS based on feature maps extracted by a pretrained convolutional neural network. Since LPIPS can better match human perceptual similarity judgments, it has been widely adopted for reconstruction-based AD method, e.g., \cite{defard2021padim}, \cite{roth2022towards} and etc. Following recent works, the difference between $\boldsymbol{y}$ and $\boldsymbol{x}_0$ at the $k$-th spatial position is calculated as:
\begin{equation}
	\begin{split}
		\left[\mathcal{D}(\boldsymbol{x}_0,\boldsymbol{y})\right]_k &= \underbrace{\eta \Vert \boldsymbol{y}_k-\left(\boldsymbol{x}_0\right)_k \Vert_1}_{\text{Pixel-level Metric}} \\
		&+ \underbrace{\sum_{i \in \mathcal{J}} \left(1- \frac{\left(\mathcal{F}_i^{(k)}(\boldsymbol{x}_0)\right)^{T} \mathcal{F}_i^{(k)}(\boldsymbol{y})}{\Vert\mathcal{F}_i^{(k)}(\boldsymbol{x}_0)\Vert\Vert\mathcal{F}_i^{(k)}(\boldsymbol{y})\Vert} \right)}_{\text{Perceptual-level Metric}} 
	\end{split}
	\label{eq:difference}
\end{equation}
in which $\mathcal{L}_1$-norm and LPIPS are incorporated to measure the differences of two images from pixel-level and perceptual-level perspectives, $\mathcal{F}_i^{(k)}$ denotes the $i$-th stage output feature of ResNet-101 proposed by \cite{zagoruyko2016wide} pretrained on ImageNet at the $k$-th spatial position, and $\eta$ is a hyperparameter to balance the weight between the pixel-level and perceptual-level metrics. $\mathcal{J}$ denotes the set of different stages of ResNet. 
In our paper, $\mathcal{J}$ is set to $\{1,2,3\}$. Since our model will generate multiple normal samples given a single test image, the pixel-wise anomaly score map for $\boldsymbol{y}$ is defined based on Eqn.(\ref{eq:difference}) as:
\begin{equation}
	\mathcal{\overline{D}} = \frac{1}{N_s}\sum_{j=1}^{N_s}\mathcal{D}(\boldsymbol{x}_0^{(j)},\boldsymbol{y})
	\label{eq:pixelScore}
\end{equation}
where $\boldsymbol{x}_0^{(j)}$ denotes the $j$-th reconstructed normal sample, and $N_s$ denotes the number of normal samples.

Then, based on Eqn.(\ref{eq:pixelScore}), we further design the image-wise anomaly score for $\boldsymbol{y}$ as the average of the largest $S$ pixel-wise anomaly scores in $\mathcal{\overline{D}}$ to mitigate false positives caused by image noise. In our paper, $S$ is set to 500. 

Using the above image-wise and pixel-wise anomaly scores, we design a mask image generation scheme for our observation model of Eqn.(\ref{eq:observation}). To obtain an accurate mask $\boldsymbol{m}$, we firstly set $\boldsymbol{m}$ to $\boldsymbol{1}$, i.e., all pixels in the test image $\boldsymbol{y}$ are regarded as potential anomaly pixels, and run a group of posterior sampling for normal images. Then, we obtain the score map $\mathcal{\overline{D}}$ for $\boldsymbol{y}$ based on Eqn.(\ref{eq:pixelScore}), and estimate $\boldsymbol{m}$ as:
\begin{equation}
	\boldsymbol{m}=\left(\mathcal{\overline{D}}>T_{th}\right),~T_{th} = \min\mathcal{\overline{D}} + \lambda ( \max\mathcal{\overline{D}} - \min\mathcal{\overline{D}} )
\end{equation}
where $\lambda$ is a hyperparameter, and $T_{th}$ is a threshold for anomaly score determined by $\lambda$. Using the above mask generation scheme, we re-run another group of posterior sampling for normal images and calculate anomaly scores for each image and each pixel respectively. The proposed MDPS is summarized in \textbf{Algorithm} \ref{alg:MDPS}.

\section{Experiments}

In this section, we compare our MDPS with other UAD methods, and conduct ablation studies to validate the designs.

\paragraph{Datasets.}
We conduct all experiments on the MVTec and BTAD Datasets. The MVTec dataset is an industrial AD benchmark  \cite{bergmann2021mvtec}, which contains 15 categories (5 textural categories and 10 object categories) with about 200 normal samples and 100 anomaly samples for each class. It provides various types of anomaly with pixel-level segmentation ground truths such as scratches, cracks, color, and missing components, posing a great challenge to AD. In our experiment, following \cite{roth2022towards}, we resize all images to $256 \times 256$ and center crop the images to $224 \times 224$ for MVTec.
The BTAD dataset contains approximately 2500 real-world industrial images of three products \cite{mishra2021vt}, which is more challenging for AD. Since some anomalies are located in the edge regions of the image in BTAD, we only resize all images to $256 \times 256$ without center cropping.

\begin{figure}[t]
	\centering
	\setlength{\abovecaptionskip}{5pt}
	\includegraphics[width=1\linewidth]{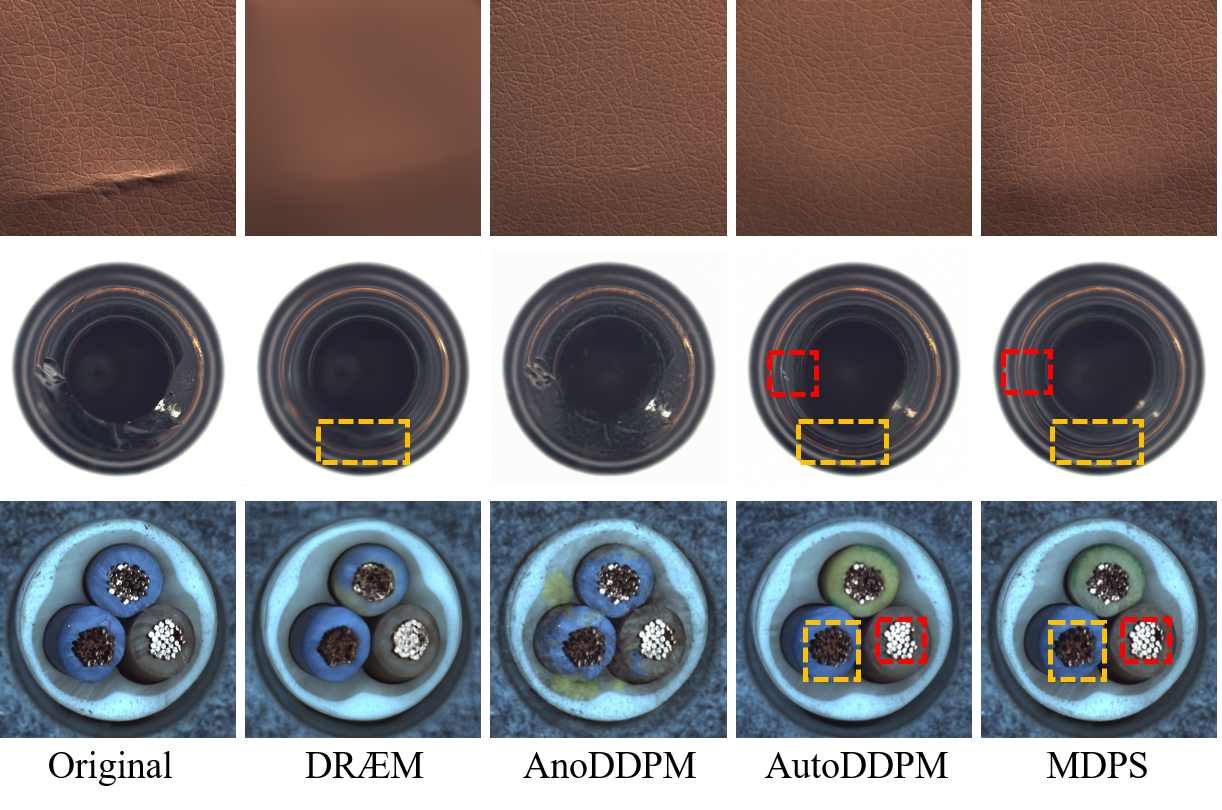}
	\caption{Comparisons of the reconstruction results MVTec. Noting the area in dotted boxes, MDPS gives the best reconstruction results.}
	\label{fig:comparison}
	\vspace{-0.1cm}
\end{figure}

\begin{table}[t]
	\centering
	\small
	\setlength\tabcolsep{0.035cm}
	\renewcommand\arraystretch{1.2}
	\setlength{\abovecaptionskip}{0cm}
	\caption{Anomaly detection and localization performance on BTAD. (Image-AUROC \%, Pixel-AUROC \%). }
	\begin{tabular}{ccccc}
		\hline
		& \multicolumn{3}{c}{\textbf{classes}}                                                                                  &                                       \\ \cline{2-4}
		\multirow{-2}{*}{\textbf{Methods}} & 01                                    & 02                                    & 03                                    & \multirow{-2}{*}{\textbf{Average}}        \\ \hline
		DRAEM & (98.5,91.5)  & (68.6,73.4)         & (99.8,96.3)          & (89.0,87.1)        \\ \hline
		VT-ADL & (97.6, \textbf{99.0})          & (71.0, 94.0)          & (82.6, 77.0)          & (83.7, 90.0)          \\ \hline
		
		AnoDDPM & (71.0,62.3)  & (60.1,60.4)          & (52.0, 53.3)          & (61.0,58.7)          \\ \hline
		AutoDDPM & (96.1,67.5)  & (76.7,59.7)          & (99.3, 74.3)          & (90.7,67.2)          \\ \hline
		
		PatchCore & (90.9, 95.5)          & (79.3, 94.7)          & (99.8, 99.3)          & (90.0, 96.5)          \\ \hline
		PaDiM & (99.8, 97.0)          & (82.0, \textbf{96.0})          & (99.4, 98.8)          & (93.7, 97.3)          \\ \hline
		PyramidFlow & (\textbf{100}, 97.4)  & (88.2, \textbf{97.6})          & (99.3, 98.1)          & (95.8, \textbf{97.7})          \\ \hline
		
		MDPS($N_s=1$)	& (\textbf{100}, 98.3)  & (\textbf{99.9}, 95.1)          & (\textbf{100}, \textbf{99.4})  & (\textbf{99.9}, 97.6)         \\ \hline
		MDPS($N_s=16$)	& (\textbf{100}, 98.4)  & (95.2, 95.3)         & (\textbf{100}, \textbf{99.4})   &(98.4, \textbf{97.7})      \\ \hline
		
	\end{tabular}
	\vspace{-0.2cm}
	\label{table_BTAD}
\end{table}

\begin{table}[t]
	\centering
	\tabcolsep=0.08cm
	\renewcommand\arraystretch{1.1}
	\setlength{\abovecaptionskip}{0pt}
	\caption{Comparison on time consumption for per image. }
	\label{tab:my-table}
	\begin{tabular}{l|cccc}
		\hline
		Method   &       Patchcore & DRÆM & AutoDDPM & MDPS($N_s=1$) \\ \hline
		Times(s) & 0.17-0.6 & 0.13 & 33.5     & 0.5  \\ \hline
	\end{tabular}
	\label{tbl:time}
	\vspace{-0.2cm}
\end{table}

\paragraph{Evaluation Metrics.}
To evaluate the results of all AD comparison methods, we employ the metrics of \textit{Area Under the Receiver Operating characteristic Curve} (AUROC). Specifically, we adopt Image-AUROC to evaluate the accuracy of anomaly detection, and adopt Pixel-AUROC to evaluate the accuracy of anomaly localization. 

\paragraph{Implementation details.}
We adopt the U-net architecture proposed by \cite{dhariwal2021diffusion} to implement the denoiser $\boldsymbol\epsilon_{\theta}(\boldsymbol{x}_t,t)$ of DDIM for sampling of $p(\boldsymbol{x}_0)$. For each category of normal samples in MVTec/BTAD, we train a UNet denoiser $\boldsymbol\epsilon_{\theta}(\boldsymbol{x}_t,t)$ separately within 2000 epochs using an Adam optimizer (learing rate: 1e-4, weight decay: 5e-2) based on a single GeForce RTX 3090 GPU. In the training process, the batchsize is set to 8, and the timestep of DDIM is set to be 1000. After training, we utilize the trained denoiser for the proposed MDPS, and let $T=200$, $N=10$, $\rho=100$. In Section \ref{sec:ablas}, we will further discuss the selection of hyperparameters for MDPS in details.

\subsection{Comparison with State-of-the-art}

\paragraph{MVTec.}
We compare MDPS with several representative reconstruction-based methods on the MVTec dataset, including DRÆM \cite{zavrtanik2021draem}, UniAD \cite{you2022unified}, AnoGAN \cite{schlegl2017unsupervised}, AnoSeg \cite{song2021anoseg},  AnoDDPM \cite{wyatt2022anoddpm}, AutoDDPM \cite{bercea2023mask}, and RAN \cite{lu2023removing}. The results are shown in Table \ref{MVTec_table} and Figure \ref{fig:comparison}. 

From Table \ref{MVTec_table} and Figure \ref{fig:comparison}, we can find that MDPS achieves the best performance in reconstruction quality as well as anomaly detection and localization especially when $N_s=16$. Specifically, the Image-AUROC of MDPS with $N_s=16$ outperforms the second best comparison method DRÆM by 0.8\%, and outperforms the second best DM-based method RAN by 5.7\%. The Pixel-AUROC of MDPS with $N_s=16$ outperforms the second best DM-based method RAN by 0.6\%, and shares the same values as the Pixel-AUROC of DRÆM. 
Although DRÆM shows the same anomaly localization performance as MDPS, the reconstruction results of DRÆM suffer from excessive blurring. Besides, DRÆM is a self-supervised method and performs badly for the real anomalies which differ significantly from the pseudo ones generated for training \cite{lu2023removing}. However, MDPS does not require any pseudo training samples, thus, has higher generalization ability than DRÆM.

\paragraph{BTAD.}
To further demonstrate the generalizability and superiority of our method, we compare MDPS with several recent state-of-the-art AD methods on the BTAD dataset, including four reconstruction-based methods (DRÆM ,VT-ADL \cite{mishra2021vt}, AnoDDPM and AutoDDPM),  two representation-based methods (PatchCore \cite{roth2022towards} and PaDiM \cite{defard2021padim}), and a normalizing flow based method (PyramidFlow \cite{lei2023pyramidflow}). The results are shown in Table \ref{table_BTAD}.
From Table \ref{table_BTAD}, we can find that MDPS still achieves the competitive performance for anomaly detection and localization under the metric of average AUROC.
Note that DRÆM shows a significant decrease performance on the BTAD dataset compared with the results of MVTec due to its limited generalization ability \cite{lu2023removing}.

Besides, we compare MDPS with several state-of-the-art AD methods on time consumption in Table \ref{tbl:time}, including PatchCore, DRÆM and AutoDDPM.  
Thanks to the acceleration of DDIM, MDPS shows comparable computational cost compared with AutoDDPM and PatchCore, but cannot compete with the AE-based method DRÆM.

\begin{figure}[t]
	\centering
	\setlength{\abovecaptionskip}{5pt}
	\includegraphics[width=1\linewidth]{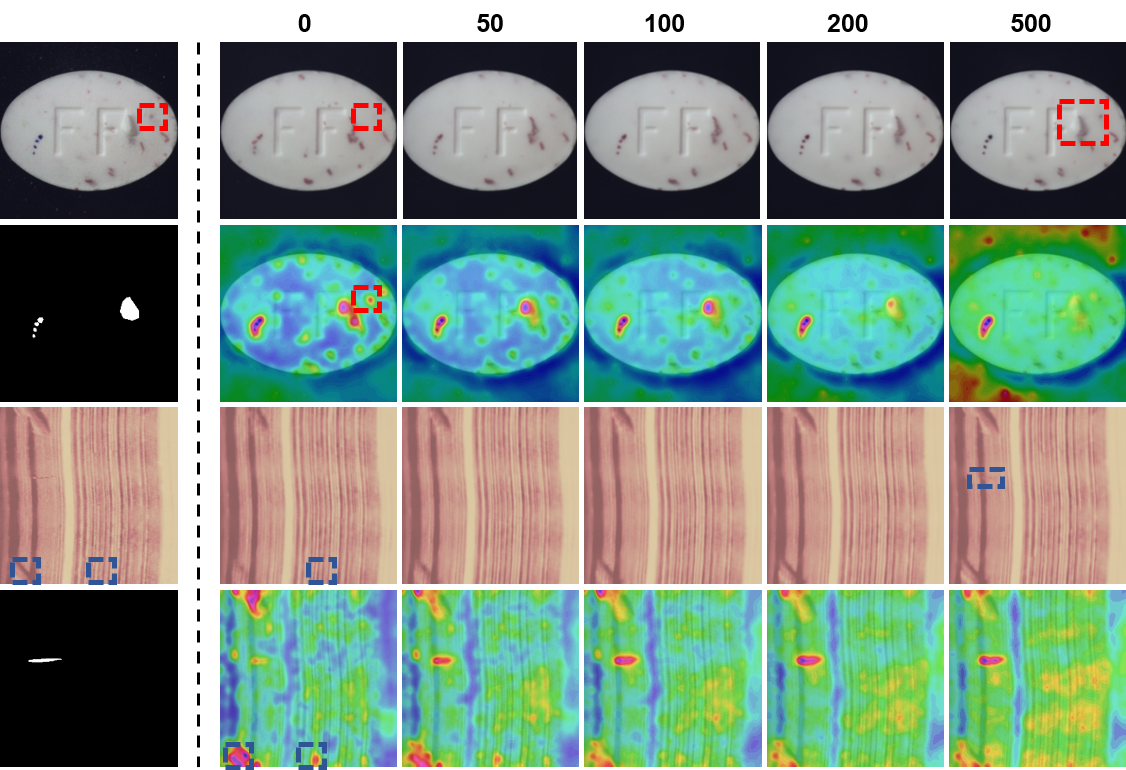}
	\caption{
		Qualitative comparison of different guidance scale $\rho$. The left side of the dotted line represents the original images and ground truths. The first and third lines on the right side of the dotted line represent the reconstructed image, and the second and fourth lines represent the heatmap. Note areas in the dotted boxes.
	}
	\label{fig:rho}
	\vspace{-0.1cm}
\end{figure}

\begin{figure}[t]
	\centering
	\setlength{\abovecaptionskip}{5pt}
	\includegraphics[width=1\linewidth]{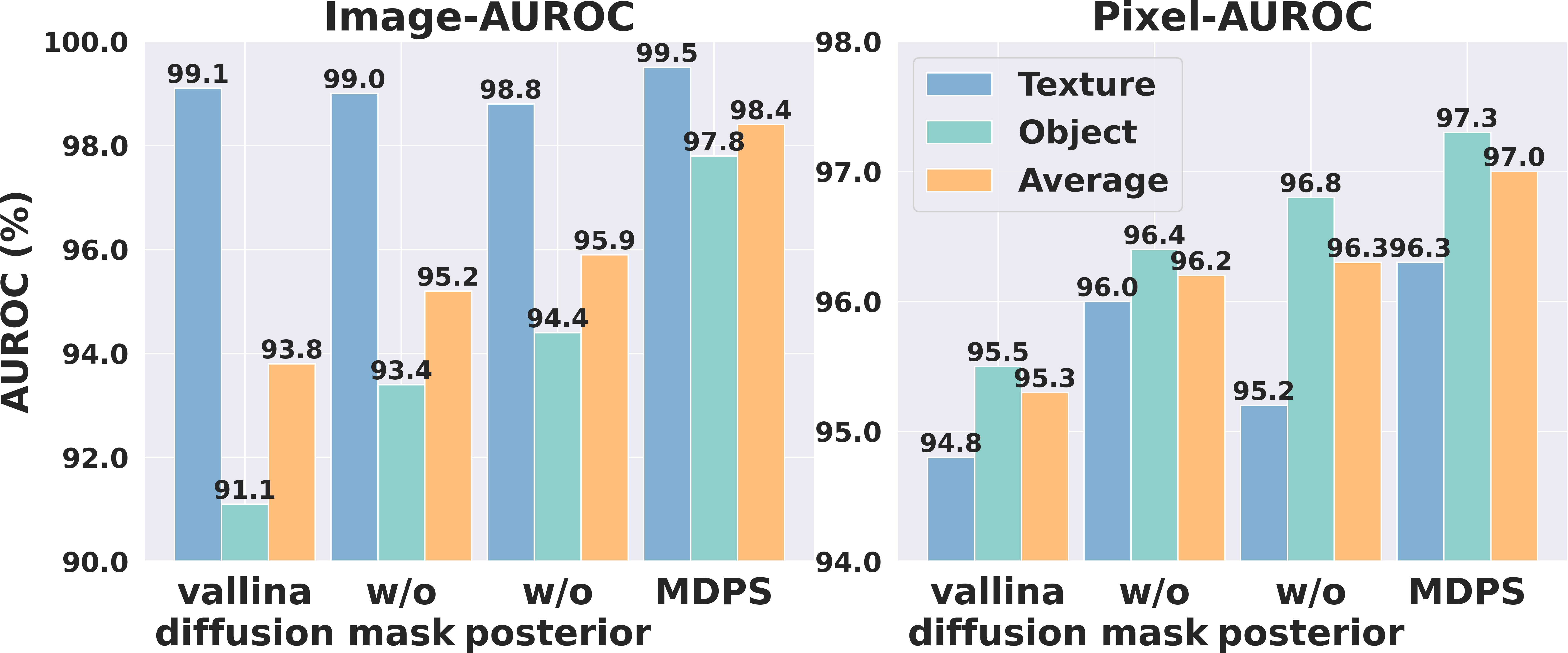}
	\caption{Ablation results of MDPS on MVtec.}
	\label{fig:ablation}
	\vspace{-0.1cm}
\end{figure}

\subsection{Ablation Study}
\label{sec:ablas}

In this section, we conduct a series of ablation experiments on MVTec to discuss the selection of hyperparameters and validate the designs of MDPS and anomaly scoring.

\begin{figure*}[!h]
	\centering
	\setlength{\abovecaptionskip}{5pt}
	\includegraphics[width=1.01\linewidth]{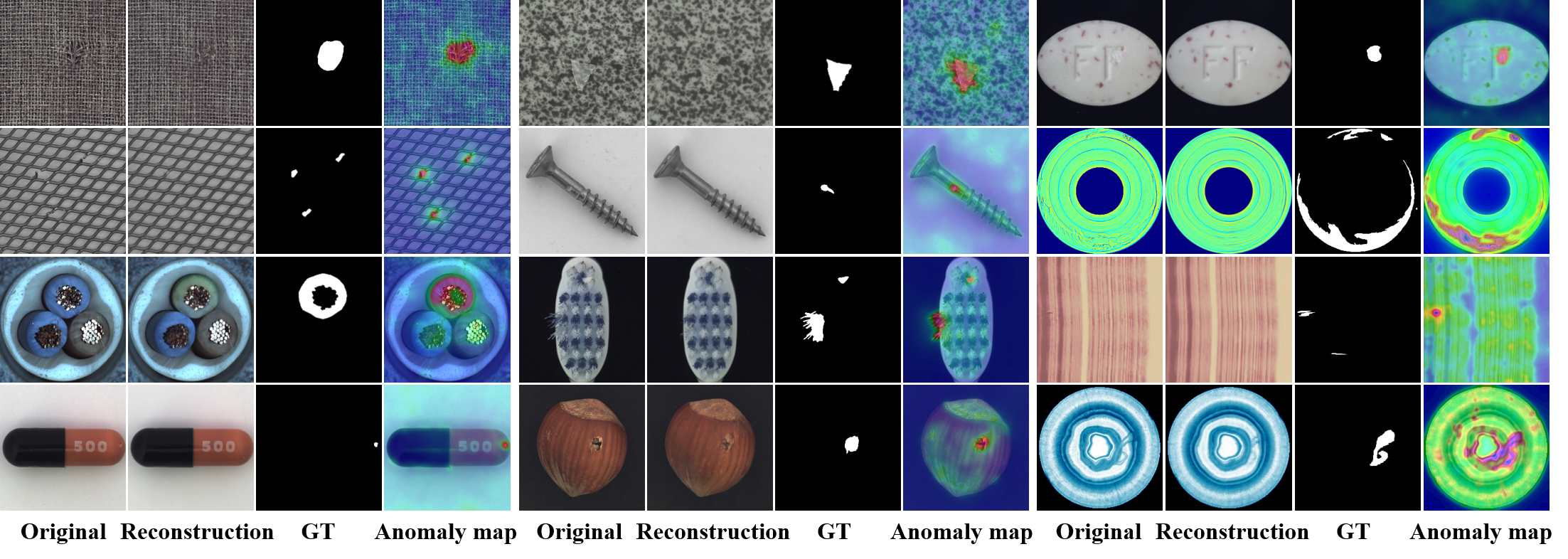}
	\caption{
		Qualitative results of our method. We choose 12 examples from MVTec and BTAD and more results can be found in the supplementary. From left to right are original images, normal reconstruction images, ground truth and our localization results.
	}
	\label{fig:qualiative}
\end{figure*}

\begin{table}[t]
	\centering
	\setlength{\abovecaptionskip}{0pt}
	\setlength\tabcolsep{0.18cm}
	\caption{Ablation results of the designed metric (AUROC \%).}
	\label{table_pixel}
	\begin{tabular}{c|ccc}
		\hline
		& Pixel-only & Perceptual-only & Pixel+Perceptual   \\ \hline
		Image & 90.7           &  91.2               & \textbf{98.8}          \\
		Pixel & 90.5           &  92.0               & \textbf{97.3} \\ \hline
	\end{tabular}
	\vspace{-0.2cm}
\end{table}
\begin{table}[t]	
	\setlength\tabcolsep{0.3cm}
	\centering
	\setlength{\abovecaptionskip}{0pt}
	\caption{The results with different $N_s$ on MVTec (AUROC \%).}
	\label{table_MS}
	\begin{tabular}{c|ccccc}
		\hline
		$N_s$    & 1     & 2     & 4     & 8     & 16            \\ \hline
		Image & 98.37 & 98.45 & 98.20 & 98.48 & \textbf{98.77} \\
		Pixel & 96.96 & 97.02 & 97.24 & 97.23 & \textbf{97.32} \\ \hline
	\end{tabular}
	\vspace{-0.2cm}
\end{table}

\paragraph{Selection of Hyperparameters.}
We empirically set $T=200$ and $N=10$ for MDPS\footnote{Selection of $T$ and $N$ are shown in Supplementary Material.}. Then, we mainly focus on selection of the guidance scale $\rho$ for Eqn.(\ref{eq:posteriorDenoiser}), since $\rho$ is related to anomalies and is a key hyperparameter to influence the reconstruction quality. Several normal image reconstruction examples of MVTec and BTAD are displayed to show the influence of $\rho$ in Figure \ref{fig:rho}. In Figure \ref{fig:rho},
partial normal texture details are destructed when $\rho<100$, which would lead to misjudgment of anomalous pixels in the normal region; however, when $\rho>100$, anomalous texture details appear in the reconstructed image, which would lead to misjudgment of normal pixels in the anomalous region. From Figure \ref{fig:rho}, we can find that $\rho$ can control the sensitivity for anomalies in the normal image reconstruction, and MDPS has better reconstruction quality when $\rho$ is set to 100.  

\paragraph{Effectiveness of MDPS.}
To validate the designs of MDPS,  we conduct a group of ablation experiments and display the results in Figure \ref{fig:ablation}. In Figure \ref{fig:ablation}, ``vanilla  DDIM''  represents only using the sampling process of DDIM, i.e, $\boldsymbol{m}=\boldsymbol{1}$ and $\rho=0$; ``w/o mask'' represents $\boldsymbol{m}=\boldsymbol{1}$; `w/o posterior' represents $\rho=0$, i.e., the problem of normal image reconstruction is modeled as prior sampling instead of posterior sampling for normal images. From Figure \ref{fig:ablation}, we can find that MDPS achieves higher values of Image-AUROC and Pixel-AUROC than MDPS with $\boldsymbol{m}=\boldsymbol{1}$ or $\rho=0$ no matter for textural or object categories, which validates the modeling of posterior sampling and the design of mask image $\boldsymbol{m}$ in MDPS.

To further validate the effectiveness of MDPS, we display qualitative results of MDPS in Figure \ref{fig:qualiative}. From Figure \ref{fig:qualiative}, we can find that MDPS can reconstruct high-quality normal images given various test images with different anomalies.

\begin{figure}[t]
	\centering
	\setlength{\abovecaptionskip}{5pt}
	\includegraphics[width=1\linewidth]{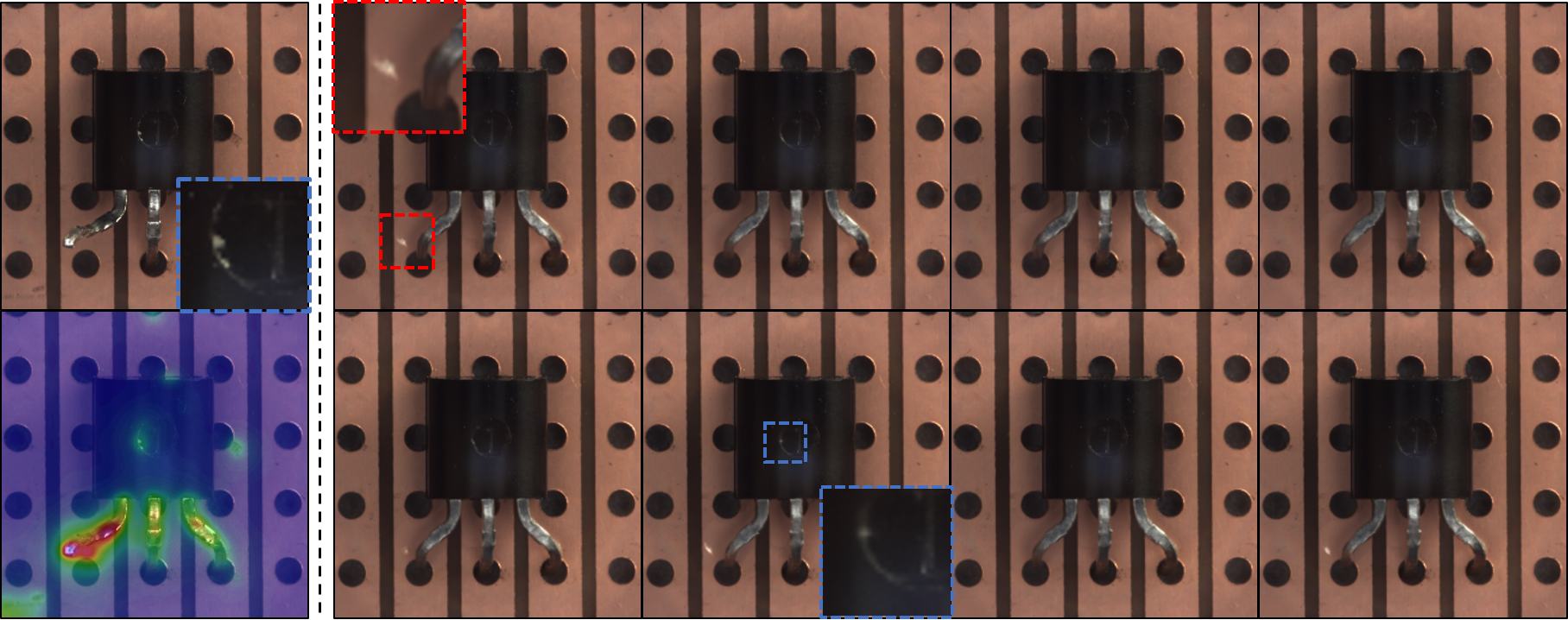}
	\caption{Multiple normal samples. \textbf{Left}: the original image and predicted heatmap. \textbf{Right}: different normal sampling results.}
	\label{figure_MNS}
	\vspace{-0.1cm}
\end{figure}

\paragraph{Effectiveness of anomaly scoring.} 
The effectiveness of anomaly scoring is related to the difference metric for two images and the number of posterior samples $N_s$. To validate the designed metric of Eqn.(\ref{eq:difference}), we compare the results of Eqn.(\ref{eq:difference}) with those of only pixel-level or perceptual-level metrics in Table \ref{table_pixel}. 
The results in Table \ref{table_pixel} show that the anomly scores obtained from pixel-level and perceptual-level can improve AUROC efficiently.

Then, we discuss the influence of $N_s$ on anomaly scoring. The results with different values of $N_s$ are shown in Table \ref{table_MS}. It can be observed that larger value of $N_s$ leads to further improvements of AUROC. In Figure \ref{figure_MNS}, we display multiple normal samplers of a test image reconstructed by MDPS. These normal images exhibit some subtle differences. But the average of difference maps for these normal samplers can rectify misjudgment caused by a single one.

\section{Conclusion}
This paper proposes MDPS, a novel and highly interpretable UAD method. MDPS generates multiple normal images based on diffusion posterior sampling under Bayesian framework. Using a combination of pixel-level and perceptual-level metrics, MDPS averages all difference maps between multiple reconstructed normal images and the test image to obtain the anomaly scores accurately. 
Exhaustive experiments show MDPS achieves high reconstruction quality and state-of-the-art performance for anomaly detection and localization compared with recent UAD methods, including 98.8\% Imgae-AUROC and 97.3\% Pixel-AUROC on MVTec dataset, 99.5\% Imgae-AUROC and 97.6\% Pixel-AUROC on BTAD dataset.
However, MDPS suffers from high computational cost caused by diffusion posterior samplings. In future work, we would try to reduce the computational cost through knowledge distillation.

\appendix

\section*{Acknowledgments}

The work is supported in part by Natural Science Foundation of Guangdong Province (2022A1515010493), in part by Shenzhen Science and Technology Program (No. JCYJ20210324140407021), in part by Natural Science Foundation of China (No. 62372082) and Natural Science Foundation of Sichuan Province (No.2023NSFSC0485).

\bibliographystyle{named}
\bibliography{ijcai24}

\end{document}